\pdfoutput=1

\documentclass[11pt]{article}

\usepackage{EMNLP2023}

\usepackage{times}
\usepackage{latexsym}

\usepackage[T1]{fontenc}

\usepackage[utf8]{inputenc}
\usepackage{microtype}

\usepackage{inconsolata}
\usepackage{graphicx}
\usepackage{amsmath}
\usepackage{amssymb}
\usepackage{algorithm}
\usepackage{algpseudocode}
\usepackage{comment}
\usepackage{graphicx}
\usepackage{multirow}
\usepackage{multicol}
\usepackage{array}
\usepackage{color, colortbl}

\newcommand{\mtrtt}[2]{\multirow{#1}{*}{\textbf{#2}}}
\newcommand{\mtctt}[2]{\multicolumn{#1}{c}{\textbf{#2}}}
\definecolor{bg}{rgb}{0.95, 0.95, 0.95}
\definecolor{mbg}{rgb}{0.75, 0.75, 0.75}
\definecolor{mmbg}{rgb}{0.65, 0.65, 0.65}
\newcommand{\STAB}[1]{\begin{tabular}{@{}c@{}}#1\end{tabular}}

\title{Okapi: Instruction-tuned Large Language Models in Multiple Languages with Reinforcement Learning from Human Feedback}

\author{{\bf Viet Dac Lai$^{1}$, Chien Van Nguyen$^{1}$, Nghia Trung Ngo$^{1}$, Thuat Nguyen$^{1}$,}\\
        {\bf Franck Dernoncourt$^2$, Ryan A. Rossi$^2$, Thien Huu Nguyen$^1$ }\\
        $^1$Dept. of Computer Science, University of Oregon, OR, USA\\
        $^2$Adobe Research, USA\\
        \texttt{\{vietl@cs,chienn,nghian@cs,thien@cs\}@uoregon.edu}\\
        \texttt{\{franck.dernoncourt,ryrossi\}@adobe.com}
        }

\begin{document}
\maketitle
\begin{abstract}


A key technology for the development of large language models (LLMs) involves instruction tuning that helps align the models' responses with human expectations to realize impressive learning abilities. Two major approaches for instruction tuning characterize supervised fine-tuning (SFT) and reinforcement learning from human feedback (RLHF), which are currently applied to produce the best commercial LLMs (e.g., ChatGPT). To improve the accessibility of LLMs for research and development efforts, various instruction-tuned open-source LLMs have also been introduced recently, e.g., Alpaca, Vicuna, to name a few. However, existing open-source LLMs have only been instruction-tuned for English and a few popular languages, thus hindering their impacts and accessibility to many other languages in the world. Among a few very recent work to explore instruction tuning for LLMs in multiple languages, SFT has been used as the only approach to instruction-tune LLMs for multiple languages. This has left a significant gap for fine-tuned LLMs based on RLHF in diverse languages and raised important questions on how RLHF can boost the performance of multilingual instruction tuning. To overcome this issue, we present Okapi, the first system with instruction-tuned LLMs based on RLHF for multiple languages. Okapi introduces instruction and response-ranked data in 26 diverse languages to facilitate the experiments and development of future multilingual LLM research. We also present benchmark datasets to enable the evaluation of generative LLMs in multiple languages. Our experiments demonstrate the advantages of RLHF for multilingual instruction over SFT for different base models and datasets. Our framework and resources are released at \url{https://github.com/nlp-uoregon/Okapi}.



\end{abstract}

\section{Introduction}


Pre-trained on massive data, large language models (LLMs) with hundreds of billions of parameters can unlock new emergent abilities that cannot be achieved with smaller models \cite{Wei2022EmergentAO}. Large generative models such as GPT-3 \cite{Rae2021ScalingLM} and OPT-175B \cite{Zhang2022OPTOP} represent some of the most recent advances in natural language processing (NLP), introducing a new learning paradigm to prompt LLMs to successfully solve a range of challenging tasks in zero-shot and few-shot fashions \cite{Kung2022Performance,Choi2023ChatGPT,Jiao2023Chatgpt,Guo2023HowCI}. However, as LLMs are trained with the autoregressive learning objective, they might exhibit unintended behaviours from human expectations \cite{Tamkin2021UnderstandingTC,Weidinger2021EthicalAS,Kenton2021AlignmentOL,Bommasani2021OnTO}. To overcome this issue, instruction fine-tuning has been proposed as a prominent approach to align LLMs with human intentions in instructions and conversations \cite{Christiano2017Deep,Stiennon2020LearningTS,Sanh2021MultitaskPT,Wei2021FinetunedLM,Ouyang2022TrainingLM}. Instruction-tuned LLMs can demonstrate significantly improved capabilities in following human instructions and avoiding the production of toxic, biased, or inaccurate texts. As such, two major techniques for instruction tuning feature supervised fine-tuning (SFT) and reinforcement learning from human feedback (RLHF) that are leveraged by the best commercial LLMs such as ChatGPT\footnote{\url{https://openai.com/blog/chatgpt/}} and GPT-4\footnote{\url{https://openai.com/research/gpt-4}} to deliver outstanding dialog performance.

Another issue with LLMs pertains to the massive scales and closed-source nature of the commercial LLMs that greatly restrict accessibility and the extent of interactions with the technology. To this end, there have been growing efforts from the open-source community to create more accessible LLMs with affordable scales while securing competitive performance as the proprietary LLMs, e.g., LLaMA \cite{Touvron2023LLaMAOA}, StableLM \cite{StableLM}, Falcon \cite{falcon40b}, and MTP \cite{MTP}. Instruction fine-tuning has also been applied to these open-source language models to improve their abilities to engage with human, and different instruction datasets have been collected either from human annotation or outputs from commercial LLMs to facilitate the tuning process, e.g., Alpaca \cite{alpaca}, Vicuna \cite{vicuna2023}, LaMini-LM \cite{Wu2023LaMiniLMAD}, and Dolly \cite{Conover2023Dolly}.

However, the instruction-following abilities of existing open-source LLMs have been developed mainly for English and some popular languages (i.e., using instruction datasets for those languages), failing to support many other languages of the world to democratize the technologies to a broader population \cite{alpaca,vicuna2023,Wu2023LaMiniLMAD}. To overcome this challenge, a few contemporary work has explored instruction tuning of multilingual LLMs for multiple languages, i.e., Phoenix \cite{Chen2023PhoenixDC} and Bactrian-X \cite{Li2023BactrianXA}. However, their multilingual instruction tuning efforts are limited to only supervised fine-tuning (SFT) techniques, which is unable to examine reinforcement learning with human feedback (RLHF) to further boost the performance for multilingual LLMs.

To fill in this gap, our work aims to develop Okapi, a open-source framework with RLHF-based instruction-tuned LLMs for multiple languages to shed light on their performance compared to the SFT methods in the multilingual settings. Okapi will emphasize on less studied languages and open-source LLMs to better democratize the benefits of instruction-tuned LLMs and provide resources for future research in this area. In particular, an example in the instruction datasets involves an instruction, an input text, and a desired response output/demonstration. In SFT, the pre-trained LLMs are fine-tuned over the instruction triples (\textit{instruction, input, output}) via supervised learning to promote their alignment with human expectations. In RLHF, generated outputs from the SFT-tuned LLMs are first ranked to provide training signals for reward functions. Afterward, the SFT-tuned models will be further optimized via reinforcement learning utilizing rewards from the trained reward models. As such, RLHF has been successfully employed to create effective commercial LLMs (e.g., InstructGPT, ChatGPT), owning to its ability to learn beyond positive examples associated with only desired demonstrations. By leveraging the reward models, RLHF can observe lower ranking scores for less accurate demonstrations to obtain richer training signals for LLMs. To our knowledge, Okapi is the first work to perform instruction tuning with RLHF for open-source LLMs over multiple languages.



To develop Okapi, we need to overcome the scarcity of necessary instruction datasets in multiple languages to train and evaluate RLHF models. Motivated by the 52K instructions from Alpaca \cite{alpaca}, we leverage Self-Instruct \cite{Wang2023SelfInstructAL} to generate 106K additional instructions in English, introducing a larger dataset to facilitate RLHF evaluation. Afterward, we utilize ChatGPT to translate the instructions into a diverse set of 26 languages, which can handle instruction examples with programming code via appropriate prompts to enhance translation quality. In addition, we introduce a translation-based prompt for ChatGPT to produce rankings for multiple responses of the same instructions from the LLMs, which will be used to train the reward models for RLHF experiments. Finally, to measure the performance of the fine-tuned LLMs in different languages, we translate three benchmark datasets for LLMs in the widely-used HuggingFace Open LLM Leaderboard \cite{Huggingface2023llm,eval-harness} into 26 languages, i.e., ARC \cite{Clark2018ThinkYH}, HellaSwag \cite{zellers-etal-2019-hellaswag}, and MMLU \cite{Hendrycks2021MeasuringMM}, using ChatGPT. These datasets challenge LLMs on diverse aspects, e.g., science reasoning, commonsense inference, world knowledge, and problem-solving, thus providing comprehensive evaluations for our models. To summarize, our contribution in this work is as follows:


\begin{itemize}
    \item {\bf Developing RLHF-tuned LLMs in multiple languages}: We present Okapi, the first instruction-tuned LLM framework, which are RLHF-based and open-source for multiple languages. Our framework covers 26 diverse languages, including some under-studied and low-resource languages for NLP, e.g., Telugu, Ukrainian, Nepali, and Kannada. Using BLOOM \cite{Scao2022BLOOMA1} and LLaMA \cite{Touvron2023LLaMAOA} as the base pre-trained LLMs, our experiments illustrate that RLHF generally performs better than SFT for multilingual instruction tuning. Our experiments also highlight the greater challenges of low-resource languages for multilingual instruction-tuning of LLMs that should be better focused in future research.
    

    
        

    \item {\bf Resource creation for instruction-tuned LLMs in multiple languages}: To cater to our experiments with multilingual RLHF, we create instruction resources for 26 different languages, including ChatGPT prompts, instruction datasets, response ranking data, benchmark datasets, and fine-tuned LLMs. We release our data, resources, and models to contribute to the development and research of multilingual instruction-tuned LLMs in the future. The resources for our Okapi framework can be found at: \url{https://github.com/nlp-uoregon/Okapi}.
\end{itemize}

\section{Data Preparation}

A key requirement for our development of instruction-tuned LLMs with RLHF involves instruction, ranking, and evaluation datasets in multiple languages, especially for low-resource languages. To this end, we perform a comprehensive data collection process to prepare necessary data for our multilingual framework Okapi in 26 languages, divided into four major steps: English instruction generation, instruction translation, ranking data production, and evaluation data creation.

\subsection{English Instruction Generation}


An instruction example to tune LLMs often has three components: an instruction to specify the task, an input text, and an associated output text (i.e., demonstration or label) \cite{Ouyang2022TrainingLM}. As such, current public instruction datasets for LLMs mainly cover English or some popular languages, which are not suitable for our experiments. Also, we note that a few recent instruction datasets such as xP3 \cite{Muennighoff2022CrosslingualGT} and Flan \cite{Chung2022ScalingIL,Longpre2023TheFC} include multilingual data; however, their instructions are still written in English. Additionally, these datasets tend to be converted from NLP task datasets with template instructions, which cannot reflect the flexibility of human-written prompts to encourage effective instruction following in different languages \cite{Wang2023SelfInstructAL}. Consequently, our goal is to develop instruction datasets with instructions, inputs, and output texts in multiple languages to better realize general prompts from human.


To achieve this goal, our strategy is to first obtain English instructions and then translate them into other languages. The benefits of our approach concern consistent instruction content across languages to facilitate performance comparison while taking advantages of translation systems to enable examination for more languages. As such, there have been several English instruction datasets collected by the open-source community to support instruction tuning of LLMs with different approaches, e.g., Alpaca \cite{alpaca}, Dolly \cite{Conover2023Dolly}, and LaMini-LM \cite{Wu2023LaMiniLMAD}. However, to conveniently scale our data and introduce variations of general instructions, we follow the instruction generation method in Alpaca, which in turn employs the Self-Instruct procedure in \cite{Wang2023SelfInstructAL}, to produce our English dataset.

\begin{figure}
    \centering
    \resizebox{.5\textwidth}{!}{
    \includegraphics{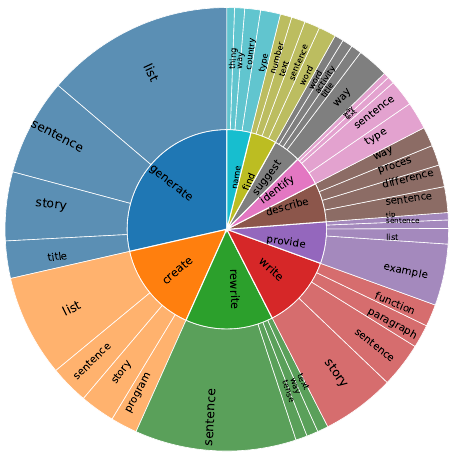}
    }
    \caption{The top 10 most frequent root verbs (inner circle) and their top 4 direct noun objects (outer circle) in the 106K generated instructions of Okapi. The instructions shown here only represent 11.4\% of all the generated instructions.}
    \label{fig:dataPlot}
\end{figure}

Starting with a pool of 175 human-written seed instructions in English over different topics, at each time, Alpaca samples several instructions from the seeds to form an in-context example to prompt the text-davinci-003 model of OpenAI for new instruction generation. The generated instructions are then compared with previous instructions using the ROUGE score, and instructions whose scores are greater than a threshold will be retained. Overall, Alpaca releases 52K instructions for tuning LLMs. In this work, we apply the same Self-Instruct procedure as Alpaca to extend its 52K instructions to a larger dataset for our RLHF-based models in Okapi. In particular, we generate 106K additional English instructions from Alpaca with two notable extensions. First, we introduce 30 new human-created instructions into the seed set from Alpaca to increase its diversity and coverage. Among others, our new instructions involve prompts for relation extraction, event extraction, event summarization, and logical questions that are not recognized in Alpaca. Second, instead of generating the new instructions from scratch, we condition our generation process on the 52K instructions from Alpaca so a new instruction is only saved if it is different enough from Alpaca's and previous instructions per the ROUGE score criteria. Figure \ref{fig:dataPlot} shows the top 10 most common root verbs and their top direct noun objects in the 106K generated instructions. These verbs and nouns represent 11.4\% of the entire set, which exhibits diverse intents and patterns in our instructions for Okapi.

\subsection{Instruction Translation}

Given the 158K English instructions from Alpaca and our generation process, we aim to translate them into multiple other languages to obtain data for our multilingual models in Okapi. Table \ref{tab:lang} presents 26 selected languages in our framework. Using the data ratios $r$ of the languages in CommonCrawl\footnote{\url{http://commoncrawl.org}} to classify languages as in previous work \cite{Bang2023AMultitask,Lai2023ChatGPTBE}, our study encompasses a diverse set of languages, including 8 high-resource languages ($r > 1.0$), 11 medium-resource languages ($r > 0.1)$, and 7 low-resource languages ($r < 0.1)$. Notably, several of our languages, such as Marathi, Gujarati, and Kannada, have received limited attention in NLP.


\begin{table}[!ht]
\centering

\resizebox{\linewidth}{!}{
\begin{tabular}{lcrrccc}
    \hline
    \mtrtt{2}{Language} & \mtrtt{2}{Code} & \mtctt{1}{Pop.} & \mtctt{2}{CC Size} & \mtrtt{2}{B} & \mtrtt{2}{L}  \\
    \cline{4-5}
     & & \multicolumn{1}{c}{(M)} & \multicolumn{1}{c}{(\%)} & Cat. & & \\
    \hline
    English     & en & 1,452    & 45.8786 & H & \checkmark & \checkmark \\
    \hline
    Russian     & ru & 258      & 5.9692 & H & \checkmark & \checkmark \\
    German      & de & 134      & 5.8811 & H & \checkmark & \checkmark \\
    Chinese     & zh & 1,118    & 4.8747 & H & \checkmark & \\
    French      & fr & 274      & 4.7254 & H & \checkmark & \checkmark \\
    Spanish     & es & 548      & 4.4690 & H & \checkmark & \checkmark \\
    Italian     & it & 68 & 2.5712 & H & \checkmark & \checkmark \\
    Dutch       & nl & 30 & 2.0585 & H & \checkmark & \checkmark \\
    Vietnamese  & vi & 85 & 1.0299 & H & \checkmark & \\
    \hline
    Indonesian  & id & 199 & 0.7991 & M & \checkmark & \\
    Arabic      & ar & 274 & 0.6658 & M & \checkmark & \\
    Hungarian & hu & 17 & 0.6093 & M & \checkmark & \checkmark \\ 
    Romanian & ro & 29 & 0.5637 & M & \checkmark & \checkmark \\ 
    Danish & da & 6 & 0.4301 & M & \checkmark & \checkmark \\ 
    Slovak & sk & 7 & 0.3777 & M & \checkmark & \checkmark \\ 
    Ukrainian   & uk & 33 & 0.3304 & M & \checkmark & \checkmark \\
    Catalan & ca & 10 & 0.2314 & M & \checkmark & \checkmark \\ 
    
    Serbian & sr & 12 & 0.2205 & M & \checkmark & \checkmark \\ 
    Croatian & hr & 14 & 0.1979 & M & \checkmark & \checkmark \\ 
    Hindi       & hi & 602 & 0.1588 & M & \checkmark & \\
    \hline
    Bengali     & bn & 272 & 0.0930 & L & \checkmark & \\
    Tamil   & ta  & 86 & 0.0446 & L & \checkmark & \\
    Nepali & ne & 25 & 0.0304 & L & \checkmark & \\
    Malayalam & ml  & 36 & 0.0222 & L & \checkmark & \\
    Marathi & mr  & 99 & 0.0213 & L & \checkmark & \\
    Telugu  & te  & 95 & 0.0183 & L & \checkmark & \\
    Kannada & kn  & 64 & 0.0122 & L & \checkmark & \\
    \hline
\end{tabular}
}
\caption{List of 26 non-English languages in our Okapi framework along with language codes, numbers of first and second speakers (the ``{\it Pop.}'' column), data ratios in the CommonCrawl corpus, and language categories. The languages are grouped into categories based on their data ratios in the CommomCrawl corpus: High Resource (H, $>1\%$), Medium Resource (M, $>0.1\%$), and Low Resource (L, $>0.01\%$) \cite{Bang2023AMultitask}. Columns ``{\it B}'' and ``{\it L}'' indicate if a language is supported by the multilingual LLMs BLOOM and LLaMa (respectively) or not.}
\label{tab:lang}
\end{table}

\begin{figure}
\centering
\fcolorbox{bg}{bg}{
\resizebox{0.9\linewidth}{!}{
\begin{minipage}{18.7em}
    \textbf{Translation Prompt:}
    Translate the values in the following JSON object into <target language> language. You must keep the keys in the JSON object in English. If a value contains programming code, only translate the comments while preserving the code. Your translations must convey all the content in the original text and cannot involve explanations or other unnecessary information. Please ensure that the translated text is natural for native speakers with correct grammar and proper word choices. Your translation must also use exact terminology to provide accurate information even for the experts in the related fields. Your output must only contain a JSON object with translated text and cannot include explanations or other information.
\end{minipage}
}
}
\caption{Translation prompt for ChatGPT for multiple languages in Okapi. We organize our instruction examples into JSON objects with fields for translation prompts, instructions, inputs, and outputs send to ChatGPT. <target language> is replaced with the selected languages in our dataset.}
\label{fig:tranPrompt}
\end{figure}

We utilize ChatGPT to translate the 158K English instructions into 26 target languages for Okapi. Compared to traditional machine translation systems, an advantage of ChatGPT for translation is the ability to use prompts to specify different expectations for the translated texts to facilitate diverse types of instructions. For example, we can instruct ChatGPT to preserve code in the instruction examples about programming as we expect code to be the same in the instructions of different natural languages. In addition, as ChatGPT has been fine-tuned on instruction-style data, we expect that it can capture the context to better translate our instructions. Figure \ref{fig:tranPrompt} shows our prompt to translate English instruction data with ChatGPT.

It is important to note that we directly translate the instruction, input text, and associated output in each English instruction example of our data. This is in contrast to the other multilingual instruction-tuning approaches \cite{Li2023BactrianXA} that only translate instructions and input texts into a target language (using Google Translate); ChatGPT is then prompted to generate response outputs in the target language for the instructions and input texts. The intuition for our approach concerns various potential issues of ChatGPT, e.g., hallucination, bias, mathematical reasoning, and toxic content \cite{Bang2023AMultitask,Borji2023ACA}, that can be exaggerated if ChatGPT is used to produce responses in non-English languages for different types of tasks/instructions \cite{Lai2023ChatGPTBE}. The diverse nature of the possible tasks/instructions will also make it more challenging to devise appropriate solutions for these problems in multilingual settings. By generating the instructions and response outputs in English, we aim to capitalize on the greater performance of LLMs for different NLP tasks in English to avoid the exaggeration issues and achieve higher quality instructions in various dimensions. By transitioning to other languages only via the translation task with ChatGPT, we can also dedicate our effort to overcome diverse multilingual challenges for instruction tuning to the translation task, which can allow convenient and effective solutions for further improvement. Table \ref{tab:stats} presents the average lengths of translated prompts and response outputs for each language in our data. Translations from Alpaca's original instructions and our new generated data are shown separately for convenient comparison.

\begin{table}[htbp]
  \centering
  \resizebox{0.8\linewidth}{!}{
    \begin{tabular}{l|c|c|c|c}
          & \multicolumn{2}{c}{Alpaca} & \multicolumn{2}{|c}{Generated} \\ \cline{2-5}
    Language & P     & R     & P     & R \\ \hline
    English & 49.0    & 56.2  & 51.2  & 58.0 \\ \hline
    Russian & 72.1  & 131.8 & 76.8  & 134.6 \\
    German & 61.1  & 94.2  & 64.6  & 96.0 \\
    Chinese & 47.2  & 47.9  & 49.3  & 48.4 \\
    French & 51.6  & 65.5  & 54.1  & 67.0 \\
    Spanish & 51.3  & 62.7  & 53.8  & 63.8 \\
    Italian & 57.4  & 86.5  & 60.5  & 87.8 \\
    Dutch & 60.3  & 94.8  & 63.9  & 96.2 \\
    Vietnamese & 53.4  & 71.3  & 56.0    & 73.2 \\ \hline
    Indonesian & 48.6  & 54.9  & 50.8  & 56.3 \\ 
    Arabic & 50.0    & 60.5  & 52.3  & 61.6 \\
    Hungarian & 68.4  & 117.3 & 72.7  & 120.0 \\
    Romanian & 62.9  & 103.9 & 66.8  & 106.7 \\
    Danish & 59.7  & 91.7  & 63.1  & 94.2 \\
    Slovak & 65.1  & 110.2 & 69.1  & 113.3 \\
    Ukrainian & 76.9  & 149.1 & 82.2  & 152.1 \\
    Catalan & 51.8  & 65.1  & 54.3  & 66.7 \\
    Serbian & 62.4  & 102.1 & 66.2  & 104.9 \\
    Croatian & 63.0    & 102.7 & 66.7  & 104.0 \\
    Hindi & 54.7  & 69.5  & 61.2  & 71.0 \\ \hline
    Bengali & 53.8  & 65.7  & 56.7  & 67.3 \\
    Tamil & 55.3  & 65.0    & 58.0    & 67.6 \\
    Nepali & 53.8  & 65.9  & 56.5  & 67.2 \\
    Malayalam & 57.8  & 75.1  & 61.0    & 77.0 \\
    Marathi & 53.6  & 67.7  & 56.3  & 69.4 \\
    Telugu & 57.1  & 74.4  & 60.2  & 75.9 \\
    Kannada & 55.4  & 69.5  & 58.2  & 71.5 \\ \hline
    Average & 57.1  & 80.8  & 60.3  & 82.6 \\
    \end{tabular}%
}
  \caption{Average lengths of translated prompts (columns ``{\it P}'') and response outputs (columns ``{\it R}'') for each language in our Okapi framework. The lengths are computed according to the number of wordpieces produced by the tokenizer of BLOOM. We separate the numbers for the translations from the original Alpaca's data (52K instructions) and our new generated data (106K instructions).}
  \label{tab:stats}%
\end{table}%


\subsection{Ranking Data Production}
\label{sec:rankingData}

\begin{figure}
\centering
\fcolorbox{bg}{bg}{
\resizebox{0.9\linewidth}{!}{
\begin{minipage}{18.7em}
    \textbf{Translation Prompt for Ranking:}
    You will be given an instruction, an input for the instruction, and four possible responses for the instruction. The input can be empty, shown as <empty>. You need to translate the provided instruction, input, and responses into English.

    {\it Instruction:} \ldots \\
    {\it Input:} \ldots \\
    {\it Response 1:} \ldots \\
    {\it Response 2:} \ldots \\
    {\it Response 3:} \ldots \\
    {\it Response 4:} \ldots
    
\end{minipage}
}
}
\caption{ChatGPT's prompt to translate target language data into English.}
\label{fig:rankTran}
\end{figure}

\begin{figure}
\centering
\fcolorbox{bg}{bg}{
\resizebox{0.85\linewidth}{!}{
\begin{minipage}{18.7em}
    \textbf{Ranking Prompt:}
    Given the translated instruction, input, and responses, you will need to rank the responses according to three factors: correctness with respect to the instruction and input, coherence, and naturalness. \\
    You will need to provide an overall rank for each response when all the three factors are considered. The overall rank for a response must be an integer between 1 and 4 where 1 is for the best response and 4 is the worst response. You cannot assign the same rank for two different responses. \\
    The format of your output must be: for each response: "<Response r>: overall rank: <1/2/3/4>". The responses must be in original order. Do not include explanation in your output. \\ \\

    \textbf{An Example Output from ChatGPT:} \\
    Response 1: 3 \\
    Response 2: 1 \\
    Response 3: 4 \\
    Response 4: 2
    
\end{minipage}
}
}
\caption{ChatGPT's prompt to rank translated data in English.}
\label{fig:rankPrompt}
\end{figure}

To perform RLHF for a LLM in Okapi, we need to obtain ranked response outputs from the model for the same instruction and input to train a reward model. Concretely, given a LLM $M$ and a dataset $S = \{{inst_k, input_k}\}_{k=1}^N$ with $N$ pairs of instructions $inst_k$ and input texts $input_k$ for a target language, we first prompt $M$ to generate $T$ output responses $output_k = \{output_k^1, \ldots, output_k^T\}$ for each pair of instruction and input text $(inst_k, input_k)$ ($T>1$). Afterward, the responses in $output_k$ are ranked according to their fitness and quality for the instruction $inst_k$ and input text $input_k$. This ranking data $\{{inst_k, input_k, output_k}\}$ can then be leveraged to train a reward model to compute a score for each triple of an instruction, an input text, and a potential response output using contrastive learning \cite{Ouyang2022TrainingLM}.

In this work, we also employ ChatGPT to rank the response outputs for multilingual LLMs. Similar to the motivation for our translation-based approach to obtain instruction data in multiple languages, our ranking strategy first asks ChatGPT to translate the instructions and responses $\{{inst_k, input_k, output_k}\}$ in a target language into English. The ranking of the responses is then done over the translated English data to exploit the greater quality of ChatGPT for English and limit different challenges associated with multilingual ranking to the translation task. To this end, we engage with ChatGPT in a two-turn dialog to obtain ranking for each example $\{{inst_k, input_k, output_k}\}$ in the target language. The first turn is to translate the example into English using the prompt in Figure \ref{fig:rankTran} while the second turn follows up with the first turn to instruct ChatGPT to rank the English translated responses using the ranking prompt in Figure \ref{fig:rankPrompt}. Our two-turn approach allows ChatGPT to condition on the translated English data in the first turn for ranking while ensuring the same format for the ranking output in the second turn for convenient parsing. Overall, we obtain ranked response outputs for 42K instructions sampled from the 106K generated instructions for each language in Okapi.


\subsection{Evaluation Data Creation}

The HuggingFace Open LLM Leaderboard \cite{Huggingface2023llm} recently adopts a suite of tasks and datasets in the Eleuther AI Language Model Evaluation Harness framework \cite{eval-harness} to facilitate performance assessment and tracking of newly developed LLMs. We employ three datasets in this leaderboard i.e., AI2 Reasoning Challenge (ARC) \cite{Clark2018ThinkYH}, HellaSwag \cite{zellers-etal-2019-hellaswag}, and MMLU \cite{Hendrycks2021MeasuringMM}, to evaluate the model performance for our Okapi framework. All the datasets are organized as multiple-choice question-answering tasks although they focus on different types of knowledge and reasoning aspects. ARC involves 1170 grade-school science questions; HellaSwag provides 9162 commonsense inference questions that are easy for humans, but difficult for many state-of-the-art models; and MMLU assesses accuracy for 13062 questions over various branches of knowledge (STEM, humanities, social sciences, and more). Nevertheless, although the LLM community has widely adopted the leaderboard for performance examination, the datasets are only provided for English, thus unable to evaluate LLMs for the languages in our work. To this end, we translate the examples of the three datasets into 26 selected languages using ChatGPT and the translation prompt in Figure \ref{fig:tranPrompt}. The translated datasets are then reserved to evaluate the LLMs in our Okapi framework.

\section{Reinforcement Learning with Human Feedback}


We follow three steps to develop a fine-tuned LLM with RLHF for each target language in our Okapi framework: supervised fine-tuning, reward model training, and reinforcement learning.

\noindent {\bf Supervised Fine-tuning (SFT)}: Starting with a multilingual pre-trained LLM as the base, e.g., BLOOM \cite{Scao2022BLOOMA1}, we fine-tune the base model with our instruction dataset for the target language using supervised learning. In Okapi, the base model is fine-tuned for three epochs via the autoregressive objective. Our training process uses a cosine learning rate schedule with $200$ warm-up steps, an initial learning rate of 2$e$-5, a batch size of $128$, and a weight decay of $0.05$. Finally, instead of leveraging approximation techniques for efficient fine-tuning, we fine-tune the entire base LLM for all of its parameters with SFT to accurately understand the model performance for multilingual settings.

\noindent {\bf Reward Model Training}: The goal of this step is to train a reward model for the target language that will compute reward signals for the reinforcement learning frameworks to further optimize the SFT-tuned model from the previous step. For each pair of a prompt and potential response, our reward model returns a scalar value to quantify the appropriateness of the response with respect to the instruction and input text in the prompt. We exploit the response-ranked datasets in Section \ref{sec:rankingData} for this training step. Using the ranking information, an example to train our reward model for a language involves an instruction and an input text (to form a prompt $x$) along with two sampled responses $y_c$ and $y_r$ for $x$ from our datasets. Based on the ranking information, we can assume one of the responses (i.e., $y_c$) is more preferable than the other (i.e., $y_r$). In the next step, the binary ranking loss \cite{Ouyang2022TrainingLM} is employed to train our reward model, aiming to assign a higher score $r(x, y_c)$ for the preferred response $y_c$ than the score $r(x, y_r)$ for $y_r$: $L_{reward}(\theta) = -\mathbb{E}_{(x, y_c, y_r)} \left[ \log \sigma(r_\theta(x, y_c) - r_\theta(x, y_r)) \right]$. For the training process, we initialize the reward model for the target language from the SFT-tuned model from previous step. We train our reward model for 2 epochs with a batch size of 64 and a learning rate of 1$e$-5, using the AdamW optimizer.

\noindent {\bf Reinforcement Learning}: With the reward model established for the target language, the SFT model undergoes additional fine-tuning through reinforcement learning (RL) to align it with human preferences. For this purpose, we employ the Proximal Policy Optimization (PPO) algorithm \cite{Ouyang2022TrainingLM}. Specifically, our training process maximizes the mean reward of the model via the objective: $L_{RL}(\phi) = -\mathbb{E}_{x \sim D_{RL}, y \sim \pi_\phi (y|x)} \left[ r_\theta(x, y) -\beta KL(x, y) \right]$. Here, $D_{RL}$ corresponds to the prompt distribution, and $\pi_\phi(y|x)$ denotes the policy or language model with parameters $\phi$ that require optimization. $\pi_\phi(y|x)$ is initialized with the SFT-tuned model $\pi_\phi(y|x)$. Also, $KL(x, y) = D_{KL}(\pi_\phi (y|x) || \pi_0 (y|x))$ is the Kullback–Leibler divergence to penalize large deviation of $\pi_\phi$ from the initial SFT policy $\pi_0$, and $\beta$ is a penalty coefficient.

During the RL training phase, we keep the entire LLM frozen and solely train the top four layers for five epochs. We employ the AdamW optimizer with $\beta_1 = 0,9$, $\beta_2 = 0.95$, and $eps = 1e-8$. The KL coefficient $\beta$ is set to $0.05$, while the weight decay is $0.1$, and the learning rate is $1e-6$. In each PPO iteration, we work with a batch size of $32$ and a clip threshold of $0.2$ in Okapi.



\section{Experiments}


Our Okapi framework utilizes two multilingual LLMs:  BLOOM \cite{Scao2022BLOOMA1} and LLaMA \cite{Touvron2023LLaMAOA} as the base models for the fine-tuning processes. We focus on their 7B-parameter versions to facilitate the computing resources and achieve fairer comparison. For each base model and target language, we carry out both SFT-based and RLHF-based instruction-tuning for the model in the following manners:


\begin{itemize}
    \item SFT: The base model is fine-tuned over the 158K translated instructions (i.e., 52K from Alpaca and 106K from our generation) in the supervised manner.
    \item RLHF: The base model is first fine-tuned with supervised training over 52K translated instructions from Alpaca. Afterward, a reward model is trained to score generated responses for input prompts using contrastive learning over the ranked responses for the 42K translated instructions in Section \ref{sec:rankingData}. Note that the ranked responses are sampled from the SFT-tuned base model over 52K translated Alpaca instructions from previous step. Finally, given the reward model, the SFT-tuned base model is further optimized via reinforcement learning over 64K remaining translated instructions from our generation set \cite{Ouyang2022TrainingLM}.
\end{itemize}


The translated datasets ARC, HellaSwag, and MMLU are exploited to evaluate the performance of the models in Okapi. Following the HuggingFace Open LLM Leaderboard, the Eleuther AI Language Model Evaluation Harness framework \cite{eval-harness} is used to compute the model performance over the datasets for each language in our framework. As a reference, we also report the performance of the base models BLOOM and LLaMA in the experiments. Finally, for BLOOM, we further compare with BLOOMZ \cite{Muennighoff2022CrosslingualGT}, which is the fine-tuned version of BLOOM over the cross-lingual task mixture dataset xP3 with millions of multilingual instructions to achieve instruction-following ability.


\begin{table}[!ht]
  \centering
\resizebox{\linewidth}{!}{
    \begin{tabular}{c|l|c|c|c|c}
    & {\bf Language} & {\bf BLOOM} & {\bf BLOOMZ} & {\bf SFT}   & {\bf RLHF} \\ \hline
    \multirow{8}{*}{\STAB{\rotatebox[origin=c]{90}{{\bf High-Resource}}}}
    & Russian & 27.5  & 25.5  & 29.2  & 30.3 \\ \cline{2-6}
    & German & 26.3  & 25.4  & 24.9  & 25.5 \\ \cline{2-6}
    & Chinese & 37.3  & 37.0    & 37.9  & 40.0 \\ \cline{2-6}
    & French & 36.7  & 37.6  & 37.6  & 41.2 \\ \cline{2-6}
    & Spanish & 38.1  & 37.2  & 39.7  & 41.5 \\ \cline{2-6}
    & Italian & 29.0    & 27.5  & 29.3  & 31.3 \\ \cline{2-6}
    & Dutch & 23.1  & 21.5  & 24.8  & 26.1 \\ \cline{2-6}
    & Vietnamese & 33.7  & 33.5  & 35.0    & 36.2 \\ \cline{2-6}
    \rowcolor{mbg}
    & {\bf Ave Group} & 31.5  & 30.7  & 32.3  & {\bf 34.0} \\ \hline
    \multirow{11}{*}{\STAB{\rotatebox[origin=c]{90}{{\bf Medium-Resource}}}}
    & Indonesian & 36.0   & 35.9  & 37.4  & 38.8 \\ \cline{2-6}
    & Arabic & 31.4  & 31.2  & 32.1  & 33.2 \\ \cline{2-6}
    & Hungarian & 25.9  & 22.8  & 25.2  & 27.5 \\ \cline{2-6}
    & Romanian & 26.9  & 23.4  & 27.5  & 30.3 \\ \cline{2-6}
    & Danish & 24.6  & 24.6  & 23.6  & 25.2 \\ \cline{2-6}
    & Slovak & 24.9  & 22.5  & 26.2  & 27.3 \\ \cline{2-6}
    & Ukrainian & 22.8  & 23.1  & 23.6  & 25.2 \\ \cline{2-6}
    & Catalan & 34.7  & 35.8  & 35.1  & 38.9 \\ \cline{2-6}
    & Serbian & 25.1  & 23.6  & 25.6  & 27.8 \\ \cline{2-6}
    & Croatian & 23.7  & 22.8  & 22.7  & 24.1 \\ \cline{2-6}
    & Hindi & 29.2  & 28.2  & 28.5  & 29.6 \\ \cline{2-6}
    \rowcolor{mbg}
    & {\bf Ave Group} & 27.7  & 26.7  & 28.0    & {\bf 29.8} \\ \hline
    \multirow{7}{*}{\STAB{\rotatebox[origin=c]{90}{{\bf Low-Resource}}}}
    & Bengali & 26.2  & 25.5  & 26.8  & 28.9 \\ \cline{2-6}
    & Tamil & 24.2  & 25.6  & 23.7  & 25.1 \\ \cline{2-6}
    & Nepali & 22.3  & 22.7  & 23.4  & 25.7 \\ \cline{2-6}
    & Malayalam & 26.4  & 25.1  & 24.6  & 24.7 \\ \cline{2-6}
    & Marathi & 27.3  & 24.8  & 25.8  & 26.0 \\ \cline{2-6}
    & Telugu & 24.3  & 25.8  & 23.9  & 24.5 \\ \cline{2-6}
    & Kannada & 24.7  & 24.6  & 24.5  & 24.6 \\ \cline{2-6}
    \rowcolor{mbg}
    & {\bf Ave Group} & 25.1  & 24.9  & 24.7  & {\bf 25.6} \\ \hline
    \rowcolor{mmbg}
    & {\bf Average}   & 28.2  & 27.4  & 28.4  & {\bf 30.0} \\
    \end{tabular}%
}
\caption{Performance of the models on the translated ARC dataset over different languages in Okapi. BLOOM 7B is used as the base LLM.}
\label{tab:bloom-arc}%
\end{table}%

\begin{table}[!ht]
  \centering
\resizebox{\linewidth}{!}{
    \begin{tabular}{c|l|c|c|c|c}
          & {\bf Language} & {\bf BLOOM} & {\bf BLOOMZ} & {\bf SFT}   & {\bf RLHF} \\ \hline
          \multirow{8}{*}{\STAB{\rotatebox[origin=c]{90}{{\bf High-Resource}}}}
    & Russian & 32.5  & 33.1  & 32.9  & 34.2 \\ \cline{2-6}
    & German & 32.4  & 33.1  & 34.7  & 35.9 \\ \cline{2-6}
    & Chinese & 51.2  & 42.6  & 51.8  & 53.8 \\ \cline{2-6}
    & French & 56.6  & 45.7  & 55.9  & 58.7 \\ \cline{2-6}
    & Spanish & 56.7  & 48.7  & 56.1  & 59.0 \\ \cline{2-6}
    & Italian & 40.8  & 40.3  & 43.1  & 44.6 \\ \cline{2-6}
    & Dutch & 31.7  & 32.3  & 32.6  & 34.9 \\ \cline{2-6}
    & Vietnamese & 48.3  & 40.6  & 49.0    & 51.3 \\ \cline{2-6}
    \rowcolor{mbg}
    & {\bf Ave Group} & 43.8  & 39.6  & 44.5  & {\bf 46.6} \\ \hline
    \multirow{11}{*}{\STAB{\rotatebox[origin=c]{90}{{\bf Medium-Resource}}}}
    & Indonesian & 49.5  & 42.0    & 50.0    & 52.2 \\ \cline{2-6}
    & Arabic & 43.3  & 39.5  & 44.3  & 47.0 \\ \cline{2-6}
    & Hungarian & 30.1  & 29.8  & 30.8  & 32.7 \\ \cline{2-6}
    & Romanian & 31.8  & 32.3  & 33.1  & 35.2 \\ \cline{2-6}
    & Danish & 31.2  & 31.5  & 33.8  & 35.7 \\ \cline{2-6}
    & Slovak & 29.8  & 29.6  & 31.4  & 32.9 \\ \cline{2-6}
    & Ukrainian & 30.0    & 30.4  & 32.2  & 33.6 \\ \cline{2-6}
    & Catalan & 51.2  & 40.3  & 50.9  & 53.8 \\ \cline{2-6}
    & Serbian & 29.9  & 30.1  & 30.7  & 33.7 \\ \cline{2-6}
    & Croatian & 30.0    & 29.4  & 30.5  & 31.6 \\ \cline{2-6}
    & Hindi & 36.4  & 34.0    & 37.7  & 39.7 \\ \cline{2-6}
    \rowcolor{mbg}
    & {\bf Ave Group} & 35.7  & 33.5  & 36.9  & {\bf 38.9} \\ \hline
    \multirow{7}{*}{\STAB{\rotatebox[origin=c]{90}{{\bf Low-Resource}}}}
    & Bengali & 32.8  & 31.5  & 33.9  & 35.4 \\ \cline{2-6}
    & Tamil & 29.4  & 29.5  & 30.0    & 30.4 \\ \cline{2-6}
    & Nepali & 30.9  & 31.9  & 32.5  & 34.1 \\ \cline{2-6}
    & Malayalam & 28.8  & 29.8  & 29.7  & 30.2 \\ \cline{2-6}
    & Marathi & 31.0    & 31.9  & 31.7  & 32.5 \\ \cline{2-6}
    & Telugu & 29.2  & 30.7  & 30.0    & 31.7 \\ \cline{2-6}
    & Kannada & 30.3  & 30.9  & 30.7  & 32.1 \\ \cline{2-6}
    \rowcolor{mbg}
    & {\bf Ave Group} & 30.3  & 30.9  & 31.2  & {\bf 32.3} \\ \hline
    \rowcolor{mmbg}
    & {\bf Average}   & 36.8  & 34.7  & 37.7  & {\bf 39.5} \\
    \end{tabular}
}
\caption{Performance of the models on the translated HellaSwag dataset over different languages in Okapi. BLOOM 7B is used as the base LLM.}
\label{tab:bloom-hellaswag}%
\end{table}%

\begin{table}[!ht]
  \centering
\resizebox{\linewidth}{!}{
    \begin{tabular}{c|l|c|c|c|c}
          & {\bf Language} & {\bf BLOOM} & {\bf BLOOMZ} & {\bf SFT}   & {\bf RLHF} \\ \hline
          \multirow{8}{*}{\STAB{\rotatebox[origin=c]{90}{{\bf High-Resource}}}}
          & Russian & 26.2  & 25.4  & 26.5  & 26.8 \\ \cline{2-6}
          & German & 28.1  & 25.6  & 27.0    & 28.6 \\ \cline{2-6}
          & Chinese & 29.1  & 27.2  & 27.7  & 28.2 \\ \cline{2-6}
          & French & 27.4  & 27.7  & 27.7  & 28.4 \\ \cline{2-6}
          & Spanish & 28.9  & 27.1  & 27.8  & 28.1 \\ \cline{2-6}
          & Italian & 25.7  & 25.8  & 25.1  & 26.0 \\ \cline{2-6}
          & Dutch & 26.4  & 26.0    & 26.1  & 26.0 \\ \cline{2-6}
          & Vietnamese & 28.1  & 26.3  & 27.0    & 27.5 \\ \cline{2-6}
          \rowcolor{mbg}
          & {\bf Ave Group} & {\bf 27.5}  & 26.4  & 26.9  & {\bf 27.5} \\ \hline
          \multirow{11}{*}{\STAB{\rotatebox[origin=c]{90}{{\bf Medium-Resource}}}}
          & Indonesian & 26.9  & 26.3  & 26.8  & 27.5 \\ \cline{2-6}
          & Arabic & 27.5  & 24.4  & 27.4  & 27.7 \\ \cline{2-6}
          & Hungarian & 26.9  & 26.1  & 25.4  & 26.3 \\ \cline{2-6}
          & Romanian & 27.4  & 25.9  & 27.6  & 27.4 \\ \cline{2-6}
          & Danish & 27.1  & 25.2  & 27.2  & 26.9 \\ \cline{2-6}
          & Slovak & 26.1  & 26.3  & 26.4  & 26.1 \\ \cline{2-6}
          & Ukrainian & 26.6  & 25.8  & 25.9  & 26.4 \\ \cline{2-6}
          & Catalan & 28.8  & 26.0    & 26.7  & 27.6 \\ \cline{2-6}
          & Serbian & 27.2  & 25.7  & 27.5  & 27.6 \\ \cline{2-6}
          & Croatian & 26.0    & 26.1  & 26.4  & 27.7 \\ \cline{2-6}
          & Hindi & 27.5  & 25.9  & 26.8  & 26.5 \\ \cline{2-6}
          \rowcolor{mbg}
          & {\bf Ave Group} & {\bf 27.1}  & 25.8  & 26.7  & {\bf 27.1} \\ \hline
          \multirow{7}{*}{\STAB{\rotatebox[origin=c]{90}{{\bf Low-Resource}}}}
          & Bengali & 28.2  & 25.9  & 27.1  & 26.8 \\ \cline{2-6}
          & Tamil & 26.6  & 26.7  & 26.1  & 26.0 \\ \cline{2-6}
          & Nepali & 26.6  & 25.6  & 25.5  & 25.2 \\ \cline{2-6}
          & Malayalam & 26.4  & 25.2  & 25.8  & 25.8 \\ \cline{2-6}
          & Marathi & 26.3  & 26.0    & 26.1  & 26.1 \\ \cline{2-6}
          & Telugu & 26.2  & 25.7  & 25.4  & 25.9 \\ \cline{2-6}
          & Kannada & 26.7  & 26.0    & 26.6  & 26.8 \\ \cline{2-6}
          \rowcolor{mbg}
          & {\bf Ave Group} & {\bf 26.7}  & 25.9  & 26.1  & 26.1 \\ \hline
          \rowcolor{mmbg}
          & {\bf Average} & {\bf 27.1}  & 26.0    & 26.6  & 26.9 \\
    \end{tabular}
}
\caption{Performance of the models on the translated MMLU dataset over different languages in Okapi. BLOOM 7B is used as the base LLM.}
\label{tab:bloom-mmlu}%
\end{table}%

\begin{table}[!ht]
  \centering
\resizebox{0.9\linewidth}{!}{
    \begin{tabular}{c|l|c|c|c}
          & {\bf Language} & {\bf LLaMA} & {\bf SFT}   & {\bf RLHF} \\ \hline
          \multirow{6}{*}{\STAB{\rotatebox[origin=c]{90}{{\bf High-Resource}}}}
          & Russian & 32.1  & 32.8  & 37.7 \\ \cline{2-5}
          & German & 35.1  & 37.5  & 39.7 \\ \cline{2-5}
          & French & 37.3  & 38.4  & 38.8 \\ \cline{2-5}
          & Spanish & 36.8  & 38.7  & 39.3 \\ \cline{2-5}
          & Italian & 35.8  & 36.3  & 39.4 \\ \cline{2-5}
          & Dutch & 33.6  & 35.2  & 37.5 \\ \cline{2-5}
          \rowcolor{mbg}
          & {\bf Ave Group} & 35.1  & 36.5  & {\bf 38.7} \\ \hline
          \multirow{8}{*}{\STAB{\rotatebox[origin=c]{90}{{\bf Medium-Resource}}}}
          & Hungarian & 29.8  & 31.4  & 33.2 \\ \cline{2-5}
          & Romanian & 32.4  & 33.8  & 37.5 \\ \cline{2-5}
          & Danish & 32.7  & 35.1  & 36.8 \\ \cline{2-5}
          & Slovak & 29.0    & 34.3  & 37.2 \\ \cline{2-5}
          & Ukrainian & 32.9  & 35.7  & 36.4 \\ \cline{2-5}
          & Catalan & 35.1  & 36.8  & 36.9 \\ \cline{2-5}
          & Serbian & 30.8  & 33.5  & 35.8 \\ \cline{2-5}
          & Croatian & 33.0    & 33.8  & 35.9 \\ \cline{2-5}
          \rowcolor{mbg}
          & {\bf Ave Group} & 32.0    & 34.3  & {\bf 36.2} \\ \hline
          \rowcolor{mmbg}
          & {\bf Average}  & 33.3  & 35.2  & {\bf 37.3} \\
    \end{tabular}%
}
\caption{Performance of the models on the translated ARC dataset over different languages in Okapi. LLaMA 7B is used as the base LLM.}
\label{tab:llama-arc}%
\end{table}%

\begin{table}[!ht]
  \centering
\resizebox{0.9\linewidth}{!}{
    \begin{tabular}{c|l|c|c|c}
          & {\bf Language} & {\bf LLaMA} & {\bf SFT}   & {\bf RLHF} \\ \hline
          \multirow{6}{*}{\STAB{\rotatebox[origin=c]{90}{{\bf High-Resource}}}}
          & Russian & 45.7  & 46.0    & 49.1 \\ \cline{2-5}
          & German & 49.9  & 49.0    & 52.6 \\ \cline{2-5}
          & French & 55.7  & 55.6  & 56.9 \\ \cline{2-5}
          & Spanish & 56.4  & 55.7  & 56.6 \\ \cline{2-5}
          & Italian & 52.0    & 52.5  & 55.9 \\ \cline{2-5}
          & Dutch & 48.7  & 48.1  & 51.3 \\ \cline{2-5}
          \rowcolor{mbg}
          & {\bf Ave Group} & 51.4  & 51.2  & {\bf 53.7} \\ \hline
          \multirow{8}{*}{\STAB{\rotatebox[origin=c]{90}{{\bf Medium-Resource}}}}
          & Hungarian & 37.9  & 38.7  & 41.0 \\ \cline{2-5}
          & Romanian & 44.9  & 45.1  & 48.7 \\ \cline{2-5}
          & Danish & 46.7  & 47.7  & 51.7 \\ \cline{2-5}
          & Slovak & 35.9  & 39.5  & 43.6 \\ \cline{2-5}
          & Ukrainian & 44.1  & 46.9  & 47.7 \\ \cline{2-5}
          & Catalan & 49.6  & 49.2  & 49.0 \\ \cline{2-5}
          & Serbian & 41.1  & 42.6  & 45.0 \\ \cline{2-5}
          & Croatian & 41.1  & 42.4  & 45.2 \\ \cline{2-5}
          \rowcolor{mbg}
          & {\bf Ave Group} & 42.7  & 44.0    & {\bf 46.5} \\ \hline
          \rowcolor{mmbg}
          & {\bf Average}  & 46.4  & 47.1  & {\bf 49.6} \\
    \end{tabular}%
}
\caption{Performance of the models on the translated HellaSwag dataset over different languages in Okapi. LLaMA 7B is used as the base LLM.}
\label{tab:llama-hellaswag}%
\end{table}%

\begin{table}[!ht]
  \centering
\resizebox{0.9\linewidth}{!}{
    \begin{tabular}{c|l|c|c|c}
          & {\bf Language} & {\bf LLaMA} & {\bf SFT}   & {\bf RLHF} \\ \hline
          \multirow{6}{*}{\STAB{\rotatebox[origin=c]{90}{{\bf High-Resource}}}}
          & Russian & 30.2  & 30.0    & 30.6 \\ \cline{2-5}
          & German & 29.9  & 30.4  & 31.7 \\ \cline{2-5}
          & French & 30.5  & 31.0    & 30.7 \\ \cline{2-5}
          & Spanish & 30.3  & 30.4  & 30.9 \\ \cline{2-5}
          & Italian & 29.9  & 30.6  & 30.4 \\ \cline{2-5}
          & Dutch & 29.8  & 30.0    & 31.1 \\ \cline{2-5}
          \rowcolor{mbg}
          & {\bf Ave Group} & 30.1  & 30.4  & {\bf 30.9} \\ \hline
          \multirow{8}{*}{\STAB{\rotatebox[origin=c]{90}{{\bf Medium-Resource}}}}
          & Hungarian & 29.0    & 29.2  & 30.1 \\ \cline{2-5}
          & Romanian & 29.7  & 29.8  & 30.9 \\ \cline{2-5}
          & Danish & 30.0    & 30.9  & 31.8 \\ \cline{2-5}
          & Slovak & 29.4  & 29.6  & 30.2 \\ \cline{2-5}
          & Ukrainian & 29.4  & 30.8  & 31.6 \\ \cline{2-5}
          & Catalan & 30.2  & 30.3  & 30.5 \\ \cline{2-5}
          & Serbian & 29.2  & 29.7  & 30.4 \\ \cline{2-5}
          & Croatian & 29.3  & 29.2  & 30.0 \\ \cline{2-5}
          \rowcolor{mbg}
          & {\bf Ave Group} & 29.5  & 29.9  & {\bf 30.7} \\ \hline
          \rowcolor{mmbg}
          & {\bf Average}  & 29.8  & 30.1  & {\bf 30.8} \\
    \end{tabular}%
}
\caption{Performance of the models on the translated MMLU dataset over different languages in Okapi. LLaMA 7B is used as the base LLM.}
\label{tab:llama-mmlu}%
\end{table}%

\noindent {\bf Evaluation}: Tables \ref{tab:bloom-arc}, \ref{tab:bloom-hellaswag}, and \ref{tab:bloom-mmlu} present the performance of the models on the ARC, HellaSwag, and MMLU datasets (respectively) when BLOOM is used as the base model. Similarly, Tables \ref{tab:llama-arc}, \ref{tab:llama-hellaswag}, and \ref{tab:llama-mmlu} report the performance with the base model LLaMA over the three datasets. In the tables, in addition to the average scores over all languages for the models, we also include the average scores for each group of languages (i.e., rows ``{\it Ave Group}'' for high-, medium-, and low-resource languages) to facilitate the comparisons. As some of our selected languages (especially the low-resource ones) are not supported by LLaMA, our tables for the experiments with LLaMA will omit those languages (see Table \ref{tab:lang}).


The first observation from the tables is that RLHF is generally better than SFT for multilingual fine-tuning of LLMs over different tasks, base models, and language groups. The improvement of average performance over all languages can go up to 2.5\% on the HellaSwag dataset with LLaMA, thus demonstrating the advantages of RLHF over SFT for fine-tuning multilingual LLMs. It is also evident from the tables that the RLHF-tuned models can significantly improve the performance of the original base models (i.e., BLOOM and LLaMa) for almost all the language groups and tasks, which further highlights the quality of the generated instruction data and the effectiveness of RLHF.

Additionally, we observe that the average performance improvement achieved through RLHF is more substantial for the ARC and HellaSwag datasets, while it is less pronounced for the MMLU dataset. Based on the nature of the datasets, we attribute this phenomenon to the better alignment between our instruction data for fine-tuning with the necessary knowledge and reasoning skills in ARC and HellaSwag than those in MMLU. In particular, ARC and HellaSwag mainly test the abilities of the models on basic knowledge (i.e., from 3rd grade to 9th) and commonsense inference while MMLU focuses on professional knowledge in different areas (e.g., STEM, social sciences, humanities). As our instructions are generated with the seeds similar to Alpaca's styles \cite{alpaca}, they tend to emphasize on general knowledge and basic inference skills, thus more aligning with the ARC and HellaSwag datasets. To this end, the generated instructions cannot well activate/complement the language and knowledge skills related to MMLU from the LLMs to attain meaningful improvement from instruction tuning.

Comparing the performance of the models across language groups, we find that the models tend to achieve the highest performance for the high-resource languages, followed by the medium-resource and low-resource languages across different base models. The performance improvement of RLHF for low-resource languages is also the least (based on the base model BLOOM), promoting it a challenging area for further research. Interestingly, our fine-tuned BLOOM models with 158K generated instructions can significantly outperform BLOOMZ over almost all the languages for the ARC, HellaSwag, and MMLU datasets using either SFT or RLHF. For example, the average performance of RLHF is 4.8\% better than those for BLOOMZ over HellaSwag. As BLOOMZ has fine-tuned BLOOM over more than 78M multilingual instructions converted from NLP datasets \cite{Muennighoff2022CrosslingualGT}, it demonstrates the higher quality of our generated instructions for multilingual instruction tuning of LLMs.


\section{Related Work}

We consider two dimensions of related work in this study, i.e., multilingual tuning and multilingual evaluation.


\noindent {\bf Multilingual Tuning}: With the introduction of the Transformer architecture \cite{Vaswani2017Attention}, various language models have been explored to boost performance for NLP tasks, including the encoder models BERT \cite{devlin-etal-2019-bert} and RoBERTa \cite{Liu2019RoBERTaAR}, the decoder models GPT \cite{Radford2019Language,Brown2020LanguageMA}, and the encoder-decoder models BART \cite{lewis-etal-2020-bart} and T5 \cite{Raffel2020Xxploring}. These language models are often trained first over English data, and then extended to other languages in two main approaches: monolingual and multilingual models. In the monolingual approach, a language model is trained specifically for a particular language, e.g., for Spanish \cite{spanish2020robert}, Japanese \cite{japanese2020robert}, French \cite{martin-etal-2020-camembert,kamal-eddine-etal-2021-barthez}, Polish \cite{polish2020robert}, Sweddish \cite{swedish2020robert}, and Hindi \cite{hindi2020robert}. In contrast, the multilingual approach explores a single language model that is trained on multilingual texts to serve multiple languages and achieve knowledge transfer for lower-resource languages, e.g., the encoder-only models mBERT \cite{devlin-etal-2019-bert}, XLM-RoBERTa \cite{conneau-etal-2020-unsupervised}, the decoder-only models mBART \cite{liu-etal-2020-multilingual-denoising} and mT5 \cite{xue-etal-2021-mt5}, and the decoder-only models BLOOM \cite{Scao2022BLOOMA1} and LLaMA \cite{Touvron2023LLaMAOA}.

Based on the pre-trained language models (PLMs), the most advanced methods for NLP in different languages involve fine-tuning the PLMs on training data of the downstream tasks \cite{Min2023Recent}, leading to state-of-the-art performance for multilingual Sentence Splitting \cite{nguyen-etal-2021-trankit}, Dependency Parsing \cite{kondratyuk-straka-2019-75}, Question Answering \cite{huang-etal-2019-unicoder}, and Named Entity Recognition \cite{pires-etal-2019-multilingual} (among others). Additionally, fine-tuning multilingual PLMs (such as XLM-RoBERTa) has proven to be an effective technique to enable zero-shot cross-lingual transfer learning across languages for various NLP tasks. This convenient approach allows for a seamless extension of NLP models to encompass larger sets of languages \cite{wu-dredze-2019-beto,Karthikeyan:20,wu-etal-2022-learning,nguyen-etal-2021-crosslingual,guzman-nateras-etal-2022-cross}.

Instruction tuning can be considered as a special type of fine-tuning techniques for PLMs where generative PLMs (e.g., GPT) are further trained with instruction data to accomplish instruction following and response alignment with human expectations. Supervised fine-tuning (SFT) is the most common instruction tuning approach that is leveraged by all of the existing LLMs, including ChatGPT, Apaca \cite{alpaca}, Vicuna \cite{vicuna2023}, and LaMini-LM \cite{Wu2023LaMiniLMAD}. Reinforcement learning from human feedback can also be used to further improve the instruction following abilities of LLMs \cite{Wei2021FinetunedLM,Ouyang2022TrainingLM} although this technique has been less explored by current open-source LLMs due to the challenges in obtaining ranking data for the reward models. For multilingual learning, instruction tuning is only applied in the form of SFT for non-English languages using multilingual LLMs, e.g., BLOOM and LLaMA, in a few contemporary work \cite{Chen2023PhoenixDC,Li2023BactrianXA,Muennighoff2022CrosslingualGT}. RLHF has not been studied for instruction tuning for non-English languages. 

\noindent {\bf Multilingual Evaluation}: A major hurdle for research in multilingual learning pertains to the scarcity of evaluation datasets for NLP tasks in various languages that hinders model development and measurement. As such, prior research has invested substantial efforts to tackle this challenge, introducing multilingual datasets for a diversity of NLP tasks. These tasks include Universal Dependencies \cite{Nivre2016Universal}, Named Entity Recognition with CoNLL 2002 and 2003 \cite{Sang2002Introduction,Sang2003Introduction}, Natural Language Inference with XNLI \cite{conneau-etal-2018-xnli}, Information Retrieval with TyDi \cite{zhang-etal-2021-mr}, Question Answering with XQuAD \cite{artetxe-etal-2020-cross}, Summarization with XWikis \cite{perez-beltrachini-lapata-2021-models}, Event Extraction with MEE \cite{pouran-ben-veyseh-etal-2022-mee}, and many other tasks with XGLUE \cite{liang-etal-2020-xglue} and XTREME \cite{Hu2020XTREME}. However, these multilingual datasets are not specifically designed for evaluation of generative LLMs as our focus in this work.


To this end, the Eleuther AI Language Model Evaluation Harness \cite{eval-harness} provides an unified framework to evaluate generative language models over different knowledge and reasoning skills. The HuggingFace Open LLM Leaderboard \cite{Huggingface2023llm} leverages four key benchmarks from this framework, i.e., ARC \cite{Clark2018ThinkYH}, HellaSwag \cite{zellers-etal-2019-hellaswag}, and MMLU \cite{Hendrycks2021MeasuringMM}, and TruthfullQA \cite{lin-etal-2022-truthfulqa}, which have been widely adopted to measure progress of LLMs. However, these datasets are not usable for our multilingual framework as they only support the evaluation for English.



\section{Conclusion}

We present the first framework, called Okapi, on instruction tuning for LLMs in multiple language using reinforcement learning from human feedback (RLHF). To address the scarcity of necessary data for multilingual instruction tuning, we introduce instruction and response-ranked data in 26 diverse languages to enable the training of supervised fine-tuning models, reward models, and reinforcement learning frameworks for multilingual LLMs. Our experiments reveal the benefits of RLHF for multilingual fine-tuning of LLMs and the challenging problems of low-resource languages in this area for future research.



\section*{Limitations}

Despite our efforts to develop and evaluate instruction-tuned LLMs in multiple languages using reinforcement learning from human feedback, our work suffers from several limitations that can be improved in future work. First, although we have attempted to cover a diverse set of 26 languages, there are still many other languages in the world that are not considered in our work. Future work can extend our system to include more languages, especially for low-resource languages, to gain a more comprehensive understanding for the language generalization of the instruction tuning methods and better democratize the technologies. Second, our system only leverages the base models BLOOM and LLaMA with 7B parameters. While this approach can facilitate the computing infrastructure of a larger group of institutions for further research, it will be beneficial to support other types of multilingual base models, e.g., the encoder-decoder model mT5 \cite{xue-etal-2021-mt5}, and other model scales (e.g., the 13B and 65B models) to strengthen the system. Third, to obtain instruction and evaluation data for the development, we automatically generate instructions in English and translate them into multiple languages using ChatGPT. We also rely on ChatGPT to obtain response-ranked data for the reward models in RLHF. Although our approach enables the extension to multiple languages with affordable development costs, the generated and translated data might involve unexpected noise. Additionally, they might not perfectly reflect human-provided instruction data in different languages. To this end, future work can improve our system with human-generated instruction and evaluation data to further examine instruction tuning for multilingual LLMs. Finally, our evaluations only investigate the performance of the models on benchmark datasets for generative LLMs, which focus on testing diverse knowledge, reasoning skills, and truthful generation. Other important concerns of generative models such as hallucination, toxicity, and biases are not evaluated explicitly in our experiments. Future work can study these issues to better characterize instruction tuning methods in the multilingual settings.

\bibliography{anthology,custom}
\bibliographystyle{acl_natbib}

\clearpage

\appendix

\end{document}